\documentclass[letterpaper,10pt,conference]{ieeeconf}
\IEEEoverridecommandlockouts
\usepackage{amsmath,amssymb}
\usepackage{booktabs}
\usepackage{graphicx}
\usepackage{xcolor}
\usepackage{multirow}
\usepackage[caption=false,font=footnotesize]{subfig}
\usepackage{algorithm}
\usepackage{algpseudocode}
\usepackage{cite}
\usepackage{hyperref}
\hypersetup{pdfauthor={},pdftitle={PRISM: PRior-guided Imagination Sampling in world Models}}

\title{\LARGE\bf PRISM: PRior-guided Imagination Sampling in world Models}
\author{
Yuhai Wang$^{1}$,
Jiawei Xia$^{2}$,
Bowen Sun$^{3}$,
Rongxuan Zhou$^{1}$,
Xiao Hu$^{1}$, \\
Yongliang Shi$^{4}$,
Jing Du$^{3}$,
Yang Ye$^{1,*}$
\\[0.5em]
$^{1}$Northeastern University 
$^{2}$University of California, Berkeley \\
$^{3}$University of Florida 
$^{4}$Qiyuan Lab \\
$^{*}$Corresponding author
}

\makeatletter
\IEEEaftertitletext{%
\vspace{-2.0\baselineskip}%
\begin{center}
\small\textbf{Project Website:} \url{https://yuhaiw.github.io/PRISM_web/}
\end{center}
\vspace{0.2\baselineskip}%
\begin{minipage}{\textwidth}
\def\@captype{figure}%
\centering
\includegraphics[width=0.9\textwidth]{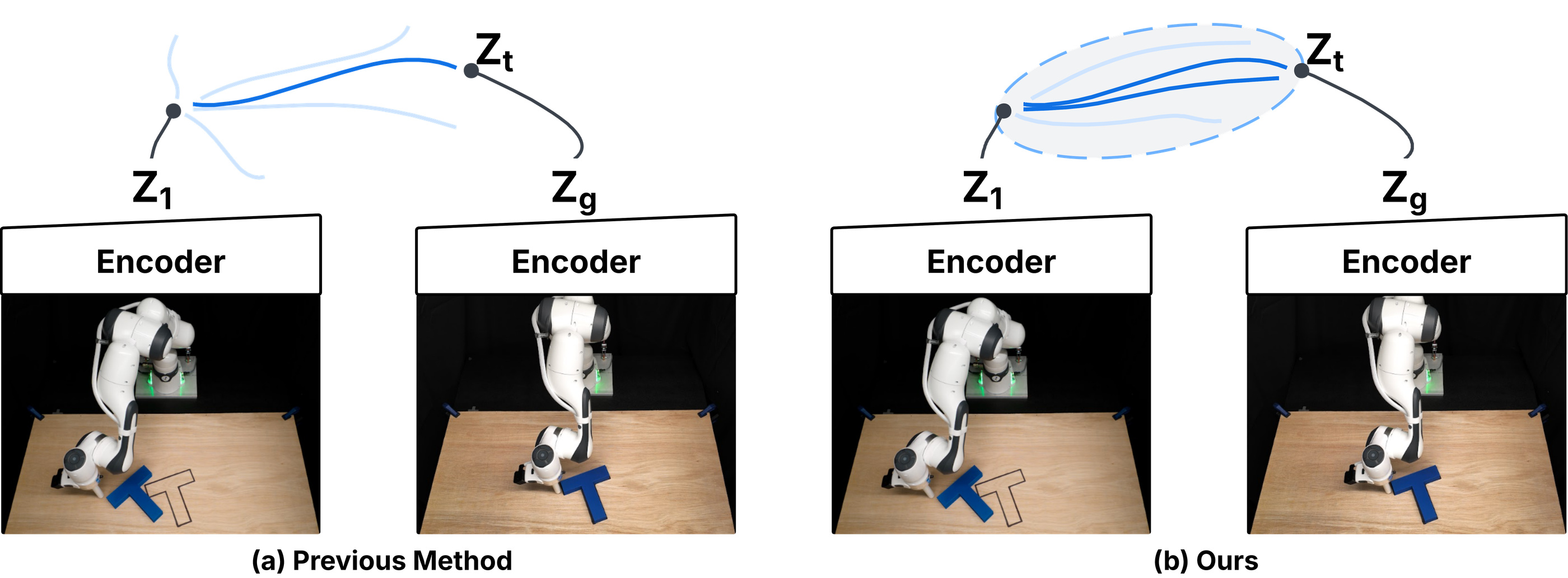}
\caption{\textbf{Planning with physical and action intuition.} Left: A planner samples candidate action sequences from an uninformed distribution and uses the world model's physical intuition to predict their outcomes. Right: PRISM also extracts action intuition from the same representations to guide sampling.}
\label{fig:teaser}
\end{minipage}
\vspace{\baselineskip}%
}
\makeatother

\newcommand{\parhead}[1]{\noindent\textbf{#1}}

\begin{document}

\maketitle
\thispagestyle{empty}
\pagestyle{empty}

\begin{abstract}

Latent world models learn representations that enable them to predict how actions change the environment—a capability we call physical intuition. We find that these same representations also support action intuition: a lightweight head can learn a goal-conditioned action prior directly from the frozen encoder’s features. We introduce PRISM to investigate how action intuition can be extracted from frozen world-model representations and used to guide goal-conditioned planning. The prior’s mean and predicted uncertainty guide the search for useful actions, while the world model evaluates their consequences against the goal. Experiments on four simulated control tasks and two real-robot platforms show that a lightweight action head improves planning performance, and that the same fusion interface accepts a diffusion prior over the same frozen features without modification. These results demonstrate the potential of reusing world-model representations to both predict action consequences and guide action selection.

\end{abstract}



\section{Introduction}
\label{sec:introduction}

Predicting how actions change the environment helps a robot evaluate their consequences before acting. Learned world models provide this capability by modeling the dynamics of an agent's surroundings from experience~\cite{ha2018world,hafner2023dreamerv3}. Joint-Embedding Predictive Architectures (JEPA) learn by predicting target representations from contextual information, without reconstructing observations~\cite{lecun2022path}. Action-conditioned world models built on this principle predict future representations from current observations and candidate actions~\cite{zhou2024dinowm,maes2026lewm}, encouraging representations that capture how the environment evolves under actions. Such representations encode a form of \emph{physical intuition}: the ability to anticipate the consequences of different actions.

To advance toward human-level intelligence, we must explore how models can use their understanding of action consequences to guide action selection. Theories of human motor control link prediction and action through paired internal models: one predicts how an action will change the body’s state, while the other determines the action needed to produce a desired change~\cite{wolpert1998paired}. This complementary ability to propose actions toward a goal is what we call \emph{action intuition}. In JEPA-based world models, predictive representations are typically used to evaluate what an action will do; their potential to help generate actions remains less fully explored. This raises a question: how much action guidance can be extracted from these predictive representations?

To investigate this question, we learn a goal-conditioned action prior from the frozen encoder's current and goal embeddings, using a simple multilayer perceptron (MLP) as a lightweight implementation. This provides a direct way to test whether the existing features support goal-conditioned action prediction without adapting the world model. We then integrate the prior into sampling-based model predictive control (MPC) through closed-form Product-of-Gaussians (PoG) fusion~\cite{hinton2002poe}. The prior's mean and predicted uncertainty guide action sampling, while the world model evaluates candidate outcomes against the goal (Fig.~\ref{fig:teaser}). Together, these extraction and integration steps form \textbf{PRISM} (\textbf{PR}ior-guided \textbf{I}magination \textbf{S}ampling in world \textbf{M}odels), a framework for probing and exploiting action intuition in world-model representations. PRISM requires no additional visual encoder, pretrained policy, or language instructions, and supports different action-head architectures.

We assess PRISM at two levels across four simulated control tasks---PushT, Cube, TwoRoom, and Reacher. First, simple MLP probes recover actions from consecutive observations encoded by the frozen world model, achieving $R^2$ values of $0.613$--$0.988$ across the four tasks. Repeating the probes with a randomly initialized, frozen encoder and projector of the same architecture yields substantially lower scores, supporting the role of learned representations in action prediction. Second, planning-based evaluations test whether the extracted priors can improve control. Across three sampling budgets and ten evaluation seeds, PRISM improves mean success over the default prior-free MPC baseline on all four tasks by $11.8$--$35.4$ percentage points. Comparisons on Franka PushT and ARX X5 cube manipulation also show higher success with PRISM. Together, these results highlight an underexplored capability of world-model representations: beyond predicting action consequences, they also support the action intuition.


\section{Related Work}
\label{sec:related}

\parhead{World models.}
World models predict the consequences of actions and support decision-making through imagined rollouts~\cite{ha2018world,hafner2023dreamerv3}. DreamerV3 learns latent dynamics with an observation-reconstruction objective and trains an actor and value function using imagined experience~\cite{hafner2023dreamerv3}. Other approaches learn predictive representations without reconstructing observations. JEPA provides a framework for prediction in representation space~\cite{lecun2022path}. DINO-WM and LeWM use latent predictions to evaluate action sequences against a visual goal~\cite{zhou2024dinowm,maes2026lewm}. We build on LeWM, whose goal-conditioned planner scores trajectories using embedding distance without reward or value heads. Keeping its world model and scoring function fixed allows us to study how an action prior changes the planner's sampling distribution.

\parhead{World models in sampling-based MPC.}
Sampling-based model predictive control (MPC) searches over action sequences by sampling candidates, predicting their outcomes, and updating the sampling distribution according to their scores. Model predictive path integral control (MPPI)~\cite{williams2017mppi} and the cross-entropy method (CEM)~\cite{deboer2005ce} are widely used approaches, with different weighting and distribution-update rules. TD-MPC2 combines a decoder-free latent world model with reward and value prediction for trajectory optimization~\cite{hansen2024tdmpc2}, whereas DINO-WM and LeWM evaluate plans through visual goal distances~\cite{zhou2024dinowm,maes2026lewm}. With a limited sampling budget, planning depends on both accurate predictions and useful action proposals. Biased-MPPI permits arbitrary sampling distributions and incorporates action sequences from ancillary controllers into its sample set~\cite{trevisan2024biasedmppi}. PRISM focuses on extracting action priors from frozen world-model representations and evaluating their utility for planning.

\parhead{Action priors for planner sampling.}
Learned action proposals can guide sampling in visual MPC. For example, VLMPC and its trajectory-generation extension use vision-language models to guide candidate action sampling~\cite{wang2024vlmpc,wang2025trajvlmpc}. PiJEPA use a language-conditioned Octo policy to initialize planning~\cite{chahe2026pijepa}. Its policy generates actions from the current observation and language instruction, while the goal image enters the planning cost. PRISM instead learns a goal-conditioned action prior directly from current and goal embeddings, without a pretrained policy or language input. Our focus is the cost and utility of extracting such priors from existing world-model features, evaluated through action-prediction probes and controlled planning experiments.


\section{Method}
\label{sec:method}

\begin{figure*}[t]
\centering
\includegraphics[width=\linewidth]{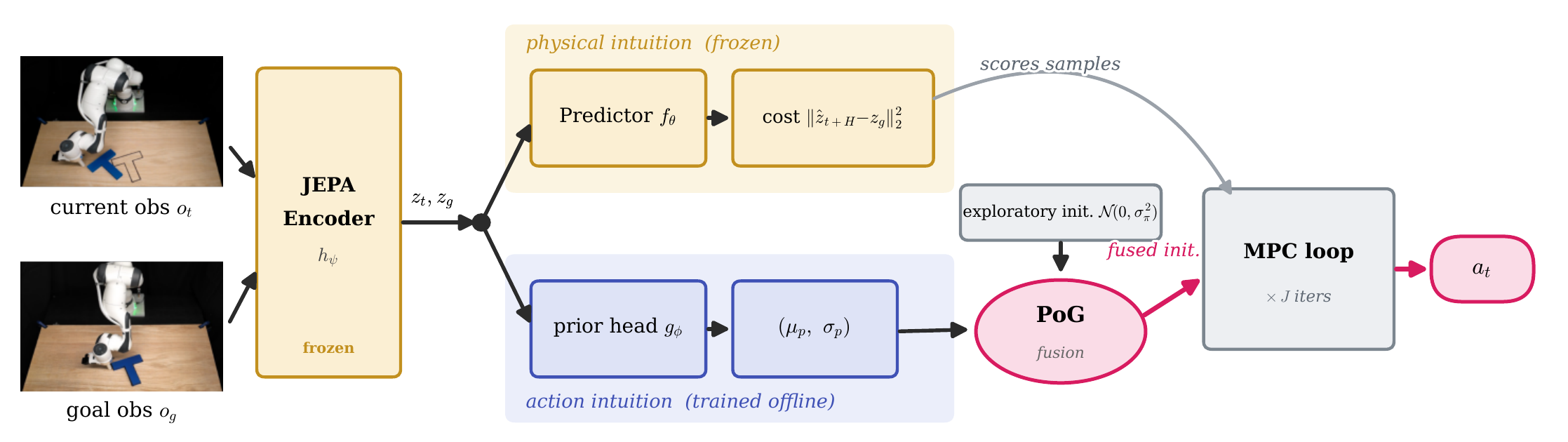}
\caption{\textbf{PRISM architecture.} The yellow group is outcome prediction, the blue group the action prior; both read the same frozen encoder, and only the prior head $g_\phi$ is trained. Grey marks the planner's default proposal, pink the fused distribution and the action it yields.}
\label{fig:architecture}
\end{figure*}

PRISM investigates whether representations learned to predict action consequences can also guide action selection. We first probe action predictability in a frozen world model, then learn an uncertainty-aware action prior from its features, and finally use this prior to guide planning (Fig.~\ref{fig:architecture}). The encoder and dynamics predictor remain fixed throughout.

\subsection{Predictive representations and action readout}
\label{sec:method-prelim}

\parhead{Frozen predictive representations.}
We build on an action-conditioned JEPA world model~\cite{maes2026lewm}, comprising an encoder $h_\psi:\mathcal O\to\mathbb R^d$ and a latent dynamics predictor $f_\theta$. The encoder maps current and goal observations to
\begin{equation}
z_t=h_\psi(o_t),\qquad z_g=h_\psi(o_g).
\end{equation}
The predictor uses the current representation and an action sequence to predict the resulting future representation. This ability to anticipate action consequences constitutes physical intuition. Our question is whether the same representations also support action intuition: predicting actions that connect a current observation to a desired outcome.

\parhead{Probing action information.}
Let $a_t$ denote the action at environment step $t$ and $\Delta$ the observation offset in environment steps. We test action predictability with a diagnostic MLP $q_\eta$ that takes two frozen embeddings and predicts the first environment-step action:
\begin{equation}
\hat a_t=q_\eta([z_t,z_{t+\Delta}]),\qquad
\mathcal L_{\mathrm{probe}}=\mathbb E\!\left[\|\hat a_t-a_t\|_2^2\right].
\end{equation}
Only $q_\eta$ is trained, using mean squared error (MSE). At $\Delta{=}1$, the probe predicts the action connecting consecutive observations, an inverse-dynamics readout. Larger offsets test whether the features support action prediction toward more distant observations. We train a probe for each task and offset $\Delta\in\{1,5,25\}$ and evaluate $R^2$ on held-out episodes (Sec.~\ref{sec:exp-probe}). These probes assess action predictability without providing planning priors.

\subsection{Extracting an uncertainty-aware action prior}
\label{sec:method-prior}

To use action intuition for control, we learn a goal-conditioned prior $p_\phi(\mathbf a\mid z_t,z_g)$ from the same task data used by the world model. The prior predicts an action-sequence distribution from current and goal embeddings, requiring no additional visual encoder or language instructions.

\parhead{Lightweight Gaussian readout.}
Our main implementation uses an MLP $g_\phi$ to predict a diagonal Gaussian:
\begin{equation}
(\mu_p,\sigma_p)=g_\phi([z_t,z_g]),\qquad
p_\phi=\mathcal N(\mu_p,\operatorname{diag}(\sigma_p^2)),
\end{equation}
where $\mu_p,\sigma_p\in\mathbb R^{HBA}$ describe $H$ planning steps, each containing $B$ actions of dimension $A$. Unlike the diagnostic probes, this head learns an action distribution for planning.

\parhead{Head architecture and training.}
The MLP uses two $512$-unit hidden layers with GELU activations and a $2HBA$-dimensional output for the mean and standard deviation, parameterized as $\sigma_p=\mathrm{softplus}(r_\phi)+0.05$, where $r_\phi$ is the raw output of the standard-deviation branch. Following Seitzer et al.~\cite{seitzer2022betanll}, we train with the $\beta$-weighted negative log-likelihood ($\beta$-NLL) loss ($\beta{=}0.5$) on cached frozen features and normalized actions. The head has approximately $0.59$M parameters on Cube and $0.51$M on PushT; optimization settings are given in Sec.~\ref{sec:head-optimization}. PRISM also supports diffusion heads conditioned on the same frozen features; the mean and variance of their sampled action sequences are used for PoG fusion. All networks remain frozen during planning.

\subsection{Using action intuition for world-model planning}
\label{sec:method-fusion}

\parhead{Uncertainty-weighted fusion.}
At each replanning step, we combine the prior with the planner's default proposal $\mathcal N(\mu_\pi,\operatorname{diag}(\sigma_\pi^2))$, where $\mu_\pi=0$. Following the product-of-experts principle~\cite{hinton2002poe}, multiplying the two diagonal Gaussians gives the per-coordinate update
\begin{align}
\sigma_\text{fused}^{2}&=(\tau_\pi+\tau_p)^{-1},\label{eq:fusion_cov}\\
\mu_\text{fused}&=\sigma_\text{fused}^{2}(\tau_\pi\mu_\pi+\tau_p\mu_p),\label{eq:fusion_mean}
\end{align}
where $\tau_\pi=\sigma_\pi^{-2}$ and $\tau_p=(s\sigma_p)^{-2}$ are elementwise precisions. The update depends only on $s\sigma_p/\sigma_\pi$; throughout we use the solver default $\sigma_\pi{=}1.0$ in normalized action space, $s{=}1$ for the MLP results, and floor $\sigma_\text{fused}$ at $0.05$.

\begin{figure*}[t]
\centering
\includegraphics[width=\textwidth]{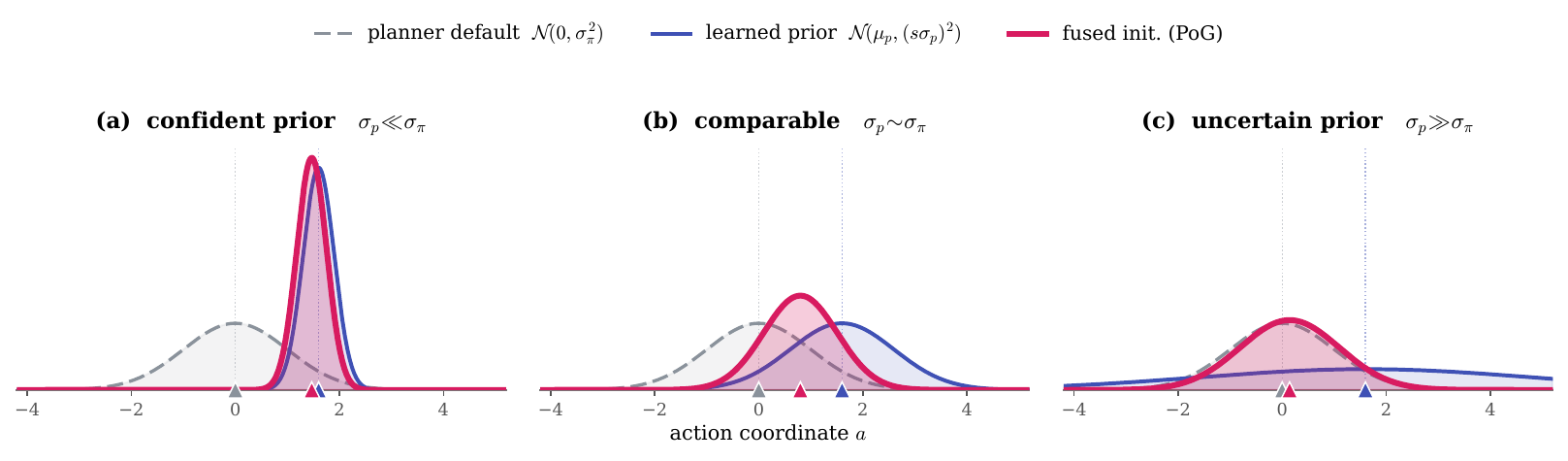}
\caption{\textbf{Uncertainty-weighted action sampling.} Grey is the default proposal, blue the action prior, pink their product; the panels vary the prior's precision relative to the planner's. Drawn before the standard-deviation floor is applied.}
\label{fig:fusion}
\end{figure*}

The prior shifts the sampling mean and narrows the sampling distribution according to its predicted uncertainty. As $s\sigma_p$ increases, its influence diminishes and the fused proposal approaches the default distribution (Fig.~\ref{fig:fusion}).

\parhead{Evaluating and refining actions.}
Model predictive control uses the fused proposal to sample candidate action sequences. The frozen world model predicts their outcomes and scores them by distance to the goal embedding:
\begin{equation}
\mathrm{cost}(\mathbf a)=\|f_\theta(z_t,\mathbf a)-z_g\|_2^2.
\label{eq:lewm_cost}
\end{equation}
Here $f_\theta(z_t,\mathbf a)$ is the predicted terminal embedding. We use MPPI~\cite{williams2017mppi} to update the sampling mean by softmax-weighting the lowest-cost candidates, while keeping the fused covariance fixed within each replanning step. The controller then executes the first action and replans from the next observation. Thus, action intuition guides which candidates are explored, while physical intuition evaluates their predicted consequences. Algorithm~\ref{alg:prism} summarizes the planning procedure.

\begin{algorithm}[htbp]
\caption{PRISM-MPPI planning step at environment time $t$}
\label{alg:prism}
\begin{algorithmic}[1]
\Require frozen $h_\psi$, predictor $f_\theta$, frozen head $g_\phi$, prior scale $s$, horizon $H$, iterations $J$, samples $K$, elite count $K_{\mathrm{elite}}$, softmax temperature $\lambda>0$
\State $z_t \gets h_\psi(o_t)$,\; $z_g \gets h_\psi(o_g)$
\State $(\mu_p, \sigma_p) \gets g_\phi([z_t,z_g])$ \Comment{action-intuition prior over $HB$ environment-step actions}
\State $(\mu_\text{fused}, \sigma_\text{fused}) \gets \mathrm{PoG}\!\big((\mu_\pi, \sigma_\pi),\,(\mu_p, s\sigma_p)\big)$ \Comment{Eqs.~\ref{eq:fusion_cov}--\ref{eq:fusion_mean}}
\State $\mu \gets \mu_\text{fused}$ \Comment{$\sigma_\text{fused}$ held fixed for all iterations}
\For{$j \gets 1$ to $J$}
  \State Draw $\{\mathbf a^{(i)}\}_{i=1}^K \sim \mathcal{N}(\mu, \mathrm{diag}(\sigma_\text{fused}^2))$
  \State $\text{cost}^{(i)} \gets \|f_\theta(z_t, \mathbf a^{(i)}) - z_g\|_2^2$
  \State Retain the $K_{\mathrm{elite}}$ lowest-cost candidates
  \State $\mu \gets \sum_i w_i\, \mathbf a^{(i)},\quad w_i \propto \exp(-\text{cost}^{(i)}/\lambda)$ \Comment{normalize weights over retained candidates}
\EndFor
\State \Return first action of $\mu$
\end{algorithmic}
\end{algorithm}


\section{Experiments}
\label{sec:experiments}

We evaluate action intuition at two levels: predicting demonstrated actions from frozen features (Sec.~\ref{sec:exp-probe}) and using an extracted action prior to improve planning (Sec.~\ref{sec:exp-main}). Both evaluations cover four goal-conditioned visual control tasks: PushT, Cube, TwoRoom, and Reacher. Additional analyses on PushT and Cube examine the effects of prior design, training-data corruption, and planner choice.

\parhead{Environments and datasets.}
PushT requires a planar pusher to align a T-block with a target pose ($A{=}2$)~\cite{chi2023diffusion}; Cube is the \texttt{cube-single} robotic arm task from OGBench ($A{=}5$)~\cite{park2024ogbench}; TwoRoom is 2D navigation through a door between two rooms ($A{=}2$); Reacher is a two-joint arm reaching a target configuration ($A{=}2$). The four datasets differ in demonstrator quality, which matters for how much action information survives to the planner's goal distance: PushT provides $18{,}685$ Diffusion-Policy episodes ($99.5\%$ trajectory success), Cube $10{,}000$ scripted-oracle episodes ($100\%$), TwoRoom $10{,}000$ episodes from a noisy two-stage heuristic ($40.4\%$), and Reacher $10{,}000$ episodes in which no episode reaches the success criterion. On each task the world model and the action head use the same dataset and the same frozen LeWM encoder~\cite{maes2026lewm}. The deployed head is trained with $\beta$-NLL as described in Sec.~\ref{sec:method-prior}; the diagnostic probe protocol is specified below.

\parhead{Head optimization.}
\label{sec:head-optimization}
The deployed heads use AdamW (learning rate $3\times10^{-4}$, weight decay $10^{-4}$), batch size $256$, and $50$ epochs, with a $1{,}000$-step linear warm-up followed by cosine decay. We split episodes $90/10$ for training and validation and select the lowest-validation-NLL checkpoint. Training windows contain $HB$ valid actions and exclude invalid terminal actions. The same StandardScaler action normalization is used during training and deployment.

\begin{figure*}[t]
\centering
\includegraphics[width=0.9\linewidth]{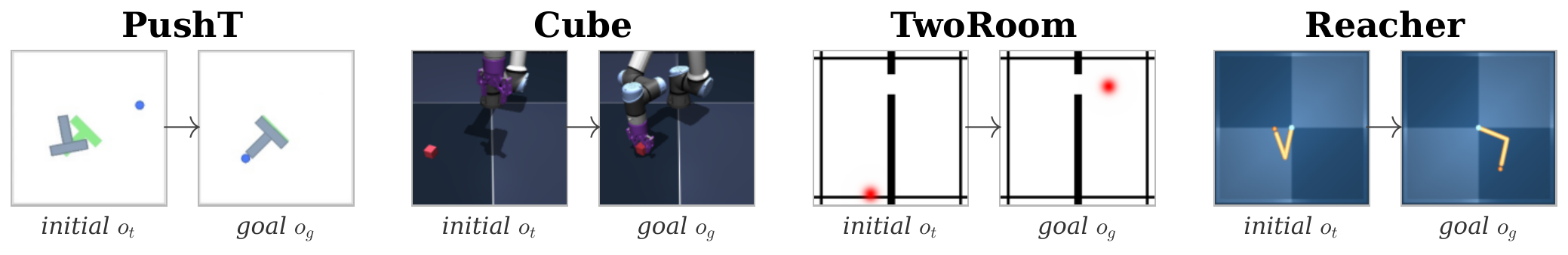}
\caption{\textbf{Simulation tasks.} One episode's first and last observation per task; at evaluation the goal is instead drawn $25$ environment steps ahead of the start. In PushT the grey T-block must reach the green pose and the blue disk is the pusher; TwoRoom renders only the agent, so its goal image is the agent at the target.}
\label{fig:tasks}
\end{figure*}

\parhead{Simulation implementation.}
The deployed-head episode splits are $16{,}817/1{,}868$ for PushT and $9{,}000/1{,}000$ for Cube, TwoRoom, and Reacher. Simulation planning uses $J{=}30$ iterations and softmax temperature $\lambda{=}0.5$, with the task-specific success criteria used by LeWM. All simulation experiments run on a single RTX~5090 with CUDA~12.8. Figure~\ref{fig:tasks} shows one observation--goal pair per task.

\subsection{Action prediction from frozen representations}
\label{sec:exp-probe}

We fit a separate MLP probe to predict $a_t$ from $[z_t,z_{t+\Delta}]$ for each task and temporal offset $\Delta\in\{1,5,25\}$. At $\Delta{=}1$, this is an inverse-dynamics prediction; $\Delta{=}25$ matches the temporal offset used for planning goals. Each probe uses a two-layer MLP with hidden width $512$ and GELU activations, trained with MSE on StandardScaler-normalized actions. Episodes are split $90/10$ by index, with $250{,}000$ training windows and $40{,}000$ held-out windows sampled per fit. We use AdamW (learning rate $3\times10^{-4}$, weight decay $10^{-4}$), batch size $512$, and $12$ epochs. We select the epoch with the lowest validation MSE and report its validation coefficient of determination, $R^2 = 1 - \mathrm{MSE}/\mathrm{Var}(a_t)$, averaged over three probe seeds $\{0,1,42\}$. Because actions are StandardScaler-normalized, $\mathrm{Var}(a_t)\approx 1$, and $R^2{=}0$ is the score of a predictor that always emits the dataset-mean action. These are validation scores on held-out episodes, not scores from an additional independent test split. Note that $R^2$ scores the conditional mean and is reduced by magnitude shrinkage: a prediction pointing in the correct direction at a fraction of the true magnitude scores low while still displacing a sampling distribution.

\begin{table}[t]
\centering
\small
\caption{\textbf{Action prediction from frozen features.} Probe $R^2$ (mean$\pm$std over three seeds; 40,000 held-out windows per fit). The lower block uses a randomly initialized, frozen encoder and projector with the same architecture and probe protocol. $R^2=0$ denotes mean-action prediction.}
\label{tab:probe}
\begin{tabular}{lccc}
\toprule
Task & $\Delta{=}1$ & $\Delta{=}5$ & $\Delta{=}25$ \\
\midrule
\multicolumn{4}{l}{\textit{Trained LeWM encoder (frozen)}} \\
PushT & $.798\pm.009$ & $.751\pm.003$ & $.367\pm.010$ \\
Cube & $.739\pm.001$ & $.699\pm.001$ & $.628\pm.001$ \\
TwoRoom & $.988\pm.000$ & $.219\pm.003$ & $.087\pm.002$ \\
Reacher & $.613\pm.001$ & $.215\pm.001$ & $.031\pm.001$ \\
\midrule
\multicolumn{4}{l}{\textit{Same architecture, re-initialized from scratch}} \\
PushT & $.000\pm.000$ & $-.000\pm.000$ & $.000\pm.001$ \\
Cube & $.090\pm.002$ & $.168\pm.002$ & $.094\pm.001$ \\
TwoRoom & $.000\pm.000$ & $.000\pm.000$ & $.002\pm.000$ \\
Reacher & $.000\pm.000$ & $.000\pm.000$ & $.000\pm.000$ \\
\bottomrule
\end{tabular}
\end{table}

Table~\ref{tab:probe} shows that a simple readout can recover actions from one-step transitions in frozen feature space on all four tasks ($R^2{=}0.613$--$0.988$). Predictability decreases with temporal distance: Cube and PushT retain stronger signals at $\Delta{=}25$, whereas TwoRoom and Reacher have much weaker long-offset predictions. The lower block of Table~\ref{tab:probe} repeats every fit with a randomly initialized encoder of the same architecture: nine of the twelve cells fall to $R^2{=}0$, and the one task that keeps a residual shows no decay in $\Delta$. What the probes read is therefore a property of the predictive training, not of any $192$-dimensional visual embedding.

\subsection{Planning with the extracted action prior}
\label{sec:exp-main}

\parhead{Evaluation protocol.}
We compare three planners on four tasks at $K\in\{32,64,128\}$: \emph{vanilla MPPI}, which initializes at zero mean and the default width $\sigma_\pi$; \emph{warm-start MPPI}, which initializes the mean at $\mu_p$ but keeps $\sigma_\pi$ and so discards the prior's $\sigma_p$; and \emph{PRISM-MPPI}, which uses both through the PoG update of Eqs.~\eqref{eq:fusion_cov}--\eqref{eq:fusion_mean}. Each configuration uses $50$ episodes per evaluation seed; we report mean$\pm$std across ten seeds $\{0,\ldots,9\}$. All tasks use a goal offset of $25$ environment steps, a planning horizon of $H{=}5$, and action blocks of $B{=}5$. $K$ is the number of candidate action sequences per iteration; all three planners use the same elite rule, $K_{\mathrm{elite}}{=}30$, and the prior scale is $s{=}1$ for PRISM. Warm-start discards $\sigma_p$, so $s$ does not enter that baseline. Rerunning both baselines at $K_{\mathrm{elite}}{=}K$ moves every cell by at most $1.4$~pp (24 cells, ten paired seeds; pooled $+0.09$~pp, $p{=}0.63$), so no comparison below turns on the elite count.

\begin{figure*}[t]
\centering
\includegraphics[width=0.9\linewidth]{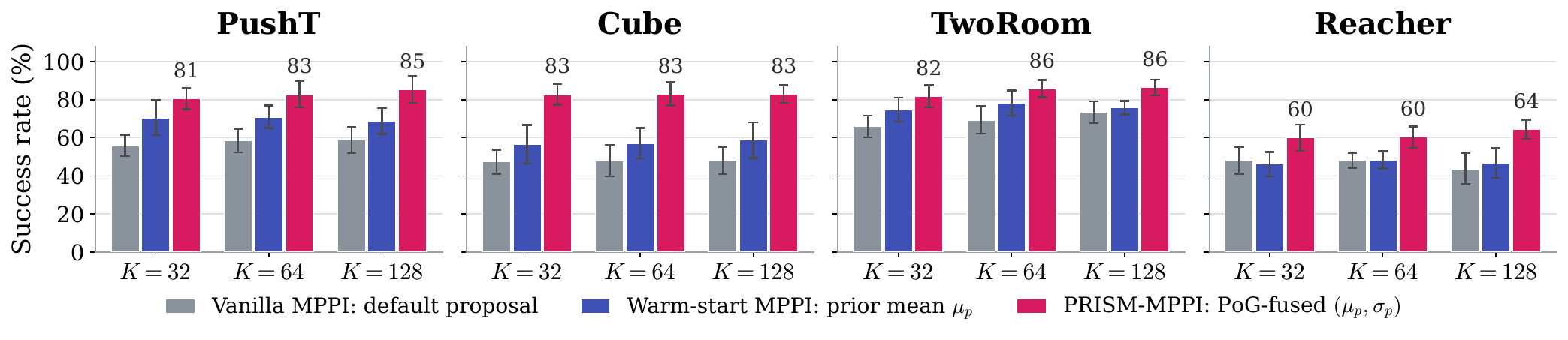}
\caption{\textbf{Four-task planning results.} Bars are means over ten evaluation seeds, whiskers one standard deviation; PRISM-MPPI means are printed above their bars. Colours carry the same meaning as in Fig.~\ref{fig:fusion}: grey the default proposal, blue the learned prior, pink the PoG-fused distribution.}
\label{fig:main}
\end{figure*}

Fig.~\ref{fig:main} shows higher mean success for PRISM than for the default prior-free baseline in all twelve task--budget settings, with gains of $11.8$--$35.4$ percentage points. PRISM also exceeds mean-only warm-starting in every setting. At $K{=}32$, PRISM outperforms vanilla MPPI at $K{=}128$ on all four tasks --- paired over the same ten seeds, $+21.8$~pp on PushT, $+34.6$~pp on Cube, $+8.4$~pp on TwoRoom and $+16.2$~pp on Reacher, winning on $10/10$ seeds in every case ($p \le 1.1\times10^{-3}$) --- so the gain comes from where sampling starts rather than from how many samples are drawn. Executing the prior's mean without candidate sampling achieves $7.0$--$75.6\%$ success across tasks and underperforms PRISM on all four (Table~\ref{tab:meanonly}), showing the benefit of using the prior within a planner.

\begin{table}[t]
\centering
\small
\caption{\textbf{Executing the prior without the planner.} SR (\%, mean$\pm$std) over ten seeds. No candidates are sampled, so there is no $K$; the PRISM column is repeated from Fig.~\ref{fig:main} for reference.}
\label{tab:meanonly}
\begin{tabular}{lcc}
\toprule
Task & Mean-only execution & PRISM-MPPI ($K{=}32$) \\
\midrule
PushT   & $28.2 \pm 5.9$ & $80.6 \pm 5.6$ \\
Cube    & $68.0 \pm 5.6$ & $82.8 \pm 5.5$ \\
TwoRoom & $75.6 \pm 4.4$ & $81.8 \pm 5.8$ \\
Reacher & $7.0 \pm 4.3$  & $60.0 \pm 6.9$ \\
\bottomrule
\end{tabular}
\end{table}

\subsection{Cost and performance of action priors}
\label{sec:exp-dphead}

We compare the MLP with an $18.5$M-parameter conditional UNet1D diffusion head based on Diffusion Policy~\cite{chi2023diffusion}, including an $18.3$M-parameter denoiser. It predicts $HB{=}25$ actions from the same frozen embeddings $[z_t;z_g]$. We estimate the prior mean and standard deviation from $16$ denoising diffusion implicit model (DDIM)~\cite{song2021ddim} samples, each using $10$ denoising steps, and apply the same PoG update and planner. The diffusion head uses goals $25$ steps ahead, while the MLP uses terminal frames; this comparison therefore does not isolate head architecture.

We also compare frozen-feature and pixel-conditioned diffusion training, holding the training windows, goal convention, schedule, and step count fixed. The pixel-conditioned policy has $19.3$M parameters and trains its own CNN. On PushT and Cube, respectively, validation denoising losses are $0.0187$ and $0.0774$ for frozen features, versus $0.0206$ and $0.0767$ for pixels. Frozen-feature head training takes approximately $23$ minutes per task, versus $254$ and $415$ minutes for the pixel-conditioned policy. The $23$-minute figure excludes feature extraction, which takes $574$ s on PushT and $573$ s on Cube and is performed once for the cached features shared by the MLP and diffusion heads. Thus, frozen features yield comparable validation denoising losses while reducing diffusion-head training time; this comparison excludes world-model pretraining.

\begin{table}[t]
\centering
\small
\setlength{\tabcolsep}{3pt}
\caption{\textbf{Cost and performance of action priors.} SR (\%, mean$\pm$std) at $K{=}32$ over ten seeds, $50$ episodes each. Both learned priors use the same world model, planner and PoG rule, but different training-goal conventions (see text). Training times on one RTX~5090 exclude world-model pretraining and feature caching; inference is prior generation per replan, excluding the shared encoder and planner.}
\label{tab:cost}
\resizebox{\columnwidth}{!}{%
\begin{tabular}{lrrrcc}
\toprule
Prior & Par.\,(M) & Train\,(min) & Infer\,(ms) & PushT & Cube \\
\midrule
None            & $0$           & --          & --           & $56.0 \pm 5.7$ & $47.4 \pm 6.3$ \\
MLP head & $0.51$--$0.59$ & $0.4$--$3.1$ & $0.048$      & $80.6 \pm 5.6$ & $82.8 \pm 5.5$ \\
Diffusion head  & $18.5$        & $23$        & $90$--$92$   & $93.6 \pm 3.6$ & $99.8 \pm 0.6$ \\
\bottomrule
\end{tabular}%
}
\end{table}

Both priors improve mean success over prior-free planning (Table~\ref{tab:cost}). The diffusion head achieves higher success, while the MLP uses $31$--$36\times$ fewer parameters and substantially less training and inference time.

End-to-end planning latency on Cube at $K{=}128$ (RTX~5090) is $210.1\pm1.2$~ms for vanilla MPPI, $210.7\pm2.2$~ms for mean-only initialization, and $211.6\pm4.8$~ms for PRISM. The measured overhead is within the run-to-run variation of the shared planning loop.

\subsection{PoG against a PiJEPA-style initialization}
\label{sec:exp-pijepa}

PiJEPA initializes the planner from the empirical mean and clamped standard deviation of its policy's samples~\cite{chahe2026pijepa}, i.e.\ it \emph{replaces} the default proposal rather than multiplying with it. This provides a direct alternative to PoG fusion, which we evaluate using the same MLP prior. PiJEPA clamps the standard deviation to $[\sigma_{\min},\sigma_{\max}]$ and reports $[0.01,0.05]$ in its own action normalization, where its prior-free baseline is initialized at $\sigma^0{=}\sigma_{\max}$; the upper clamp is therefore that planner's default sampling width, and we set $\sigma_{\max}=\sigma_\pi$ accordingly. We evaluate with the deployed head and with the head retrained on $50\%$ corrupted demonstrations (Sec.~\ref{sec:exp-corrupt}).

\begin{table}[t]
\centering
\small
\setlength{\tabcolsep}{4pt}
\caption{\textbf{Two ways to inject one prior.} SR (\%, mean$\pm$std) at $K{=}32$ over ten seeds, both rules fed the same MLP head. \emph{Weak} retrains the head on $\varepsilon{=}50\%$ corrupted demonstrations. Bold marks the better rule; Cube's weak cell is a tie in the mean.}
\label{tab:pijepa}
\begin{tabular}{llcc}
\toprule
Prior & Rule & PushT & Cube \\
\midrule
Clean & PiJEPA-style & $77.0 \pm 4.5$ & $81.2 \pm 6.4$ \\
Clean & PoG          & $\mathbf{80.6 \pm 5.6}$ & $\mathbf{82.8 \pm 5.5}$ \\
Weak  & PiJEPA-style & $45.0 \pm 7.2$ & $51.8 \pm 3.9$ \\
Weak  & PoG          & $\mathbf{62.0 \pm 5.4}$ & $51.8 \pm 6.0$ \\
\bottomrule
\end{tabular}
\end{table}

PoG is ahead in three of the four cells (Table~\ref{tab:pijepa}), and the one substantial gap is the weak prior on PushT, $+17.0$~pp on matched seeds and $10/10$ seed wins. Cube's weak cell is a tie. This difference is consistent with the larger increase in predicted uncertainty on PushT: corrupting half the demonstrations raises $\sigma_p$ from $0.53$ to $3.08$ on PushT but only from $0.60$ to $1.18$ on Cube (Table~\ref{tab:corrupt}). As predicted uncertainty increases, PoG reduces the prior's weight in the fused sampling mean. The replace rule has no such term: it installs $\mu_p$ at full strength whatever $\sigma_p$ reports. Running the same comparison with a diffusion prior instead of the head gives $+0.4$ to $+5.0$~pp, within seed noise.

\begin{table*}[t]
\centering
\caption{\textbf{Encoder ablation.} SR (\%, mean$\pm$std) over three seeds $\{0,1,42\}$ on PushT and Cube. \emph{Learned-prior params} counts the trainable action prior only; every row also uses the shared frozen world model.}
\label{tab:legacy-main}
\resizebox{\textwidth}{!}{%
\begin{tabular}{lccccccc}
\toprule
Method & Learned-prior params & \multicolumn{3}{c}{PushT} & \multicolumn{3}{c}{Cube} \\
\midrule
\multicolumn{2}{l}{\textit{Planning with World Model}} & $K{=}32$ & $K{=}64$ & $K{=}128$ & $K{=}32$ & $K{=}64$ & $K{=}128$ \\
\cmidrule(lr){3-5} \cmidrule(lr){6-8}
Vanilla MPPI (DINOv2-based) & $0$ & $4 \pm 2$ & $3 \pm 2$ & $5 \pm 1$ & $45 \pm 4$ & $49 \pm 6$ & $41 \pm 3$ \\
PRISM-MPPI (DINOv2-based) & $0.51$--$0.59$M & $13 \pm 6$ & $10 \pm 7$ & $15 \pm 5$ & $57 \pm 7$ & $65 \pm 5$ & $67 \pm 3$ \\
Vanilla MPPI (LeWM)~\cite{maes2026lewm} & $0$ & $59 \pm 5$ & $61 \pm 7$ & $57 \pm 6$ & $46 \pm 2$ & $44 \pm 5$ & $44 \pm 4$ \\
\textbf{PRISM-MPPI (ours)} & $0.51$--$0.59$M & $\mathbf{82 \pm 4}$ & $\mathbf{86 \pm 6}$ & $\mathbf{89 \pm 4}$ & $\mathbf{79 \pm 2}$ & $\mathbf{78 \pm 3}$ & $\mathbf{79 \pm 6}$ \\
\bottomrule
\end{tabular}%
}
\end{table*}

\parhead{Encoder ablation.}
Inspired by DINO-WM~\cite{zhou2024dinowm}, we evaluate our own DINOv2-based world-model implementation. Swapping the ViT-tiny encoder for frozen DINOv2-base with a learnable $768\to192$ projector, holding the LeWM predictor and training recipe fixed, PRISM keeps a $+6$ to $+18$~pp lift over vanilla MPPI ($5/6$ cells, one-sided paired $t$; Table~\ref{tab:legacy-main}, three seeds on PushT and Cube). The comparison is within-encoder: absolute levels differ between the two encoders, and the fusion does not depend on which frozen representation it reads.

\subsection{The prior transfers to CEM}
\label{sec:exp-cem}

We evaluate whether the extracted action prior also improves CEM. Using the same prior head, we compare prior-free sampling, mean-only initialization, and PoG initialization on PushT and Cube. Unlike MPPI, CEM updates both the sampling mean and covariance during optimization.

PoG initialization improves mean success over prior-free CEM at all three sampling budgets on both tasks (Table~\ref{tab:cem}). On Cube, it also outperforms mean-only initialization at every budget. On PushT, however, mean-only initialization performs better at \(K=32\) and \(64\), while the two achieve the same mean success at \(K=128\). These results show that the extracted prior can benefit CEM, although PoG does not consistently outperform mean-only initialization. On PushT at $K=32$, sweeping the initial standard-deviation floor over $\{0.05,0.10,0.20,1.0\}$ yields $24.0$--$25.3\%$ success. Increasing the initial floor therefore does not remove the gap to mean-only initialization; it does not constrain CEM's subsequent covariance refits.

\begin{table}[t]
\centering
\small
\setlength{\tabcolsep}{4pt}
\caption{\textbf{The same prior under CEM.} SR (\%, mean$\pm$std) over three seeds $\{0,1,42\}$, $50$ episodes per seed, on PushT and Cube. All three rows use CEM, which refits $\sigma$ each iteration; the MPPI counterparts are in Fig.~\ref{fig:main}, measured over ten seeds and therefore not directly comparable cell by cell.}
\label{tab:cem}
\begin{tabular}{llccc}
\toprule
Task & $K$ & No prior & Mean only & PoG \\
\midrule
PushT & $32$  & $12.7 \pm 8.1$ & $\mathbf{41.3 \pm 3.1}$ & $25.3 \pm 3.1$ \\
PushT & $64$  & $80.7 \pm 8.1$ & $\mathbf{88.7 \pm 6.1}$ & $84.0 \pm 7.2$ \\
PushT & $128$ & $90.7 \pm 6.1$ & $92.7 \pm 5.8$ & $\mathbf{92.7 \pm 2.3}$ \\
\midrule
Cube  & $32$  & $53.3 \pm 6.4$ & $60.7 \pm 1.2$ & $\mathbf{68.7 \pm 7.0}$ \\
Cube  & $64$  & $59.3 \pm 8.3$ & $74.7 \pm 2.3$ & $\mathbf{79.3 \pm 6.4}$ \\
Cube  & $128$ & $61.3 \pm 4.6$ & $76.0 \pm 4.0$ & $\mathbf{82.7 \pm 5.0}$ \\
\bottomrule
\end{tabular}
\end{table}

\subsection{Increased uncertainty reduces reliance on corrupted priors}
\label{sec:exp-corrupt}

We test how the prior responds to corrupted training data. For a fraction $\varepsilon$ of training episodes, we replace the action labels with uniform random samples within the action bounds and retrain the head. The world model and planning configuration remain fixed.

As corruption increases, both the head's mean predicted standard deviation and its held-out prediction error increase (Table~\ref{tab:corrupt}). The predicted standard deviation rises from $0.53$ to $4.49$ on PushT and from $0.60$ to $1.51$ on Cube, reducing the prior's precision and its influence on sampling. With all action labels corrupted, planning success is $60.7\%$ on PushT and $46.7\%$ on Cube, close to the respective prior-free results of $58.8\%$ and $48.2\%$. These comparisons use different seed sets, so the agreement is approximate. The results show that, under action-label corruption, the head predicts greater uncertainty and PoG correspondingly reduces reliance on the prior.

\begin{table}[t]
\centering
\small
\caption{\textbf{Degraded demonstrations.} $\sigma_p$ is the head's mean predicted standard deviation on held-out data; val MSE is its held-out error. SR at $K{=}128$ over three seeds, except the $\varepsilon{=}0$ and \emph{no prior} rows, which are the ten-seed values of Fig.~\ref{fig:main}.}
\label{tab:corrupt}
\begin{tabular}{lccc|ccc}
\toprule
& \multicolumn{3}{c|}{PushT} & \multicolumn{3}{c}{Cube} \\
$\varepsilon$ & $\sigma_p$ & val MSE & SR & $\sigma_p$ & val MSE & SR \\
\midrule
$0\%$   & $0.53$ & $0.73$ & $85.4$ & $0.60$ & $0.61$ & $83.0$ \\
$25\%$  & $1.96$ & $0.79$ & $72.0$ & $0.96$ & $0.68$ & $52.7$ \\
$50\%$  & $3.08$ & $1.07$ & $66.7$ & $1.18$ & $0.77$ & $52.7$ \\
$75\%$  & $3.90$ & $1.56$ & $66.0$ & $1.33$ & $0.90$ & $50.7$ \\
$100\%$ & $4.49$ & $2.16$ & $60.7$ & $1.51$ & $1.04$ & $46.7$ \\
\midrule
\multicolumn{1}{l}{\emph{no prior}} & --- & --- & $58.8$ & --- & --- & $48.2$ \\
\bottomrule
\end{tabular}
\end{table}

\subsection{Sensitivity to the prior scale}

\begin{figure}[htbp]
\centering
\includegraphics[width=\linewidth]{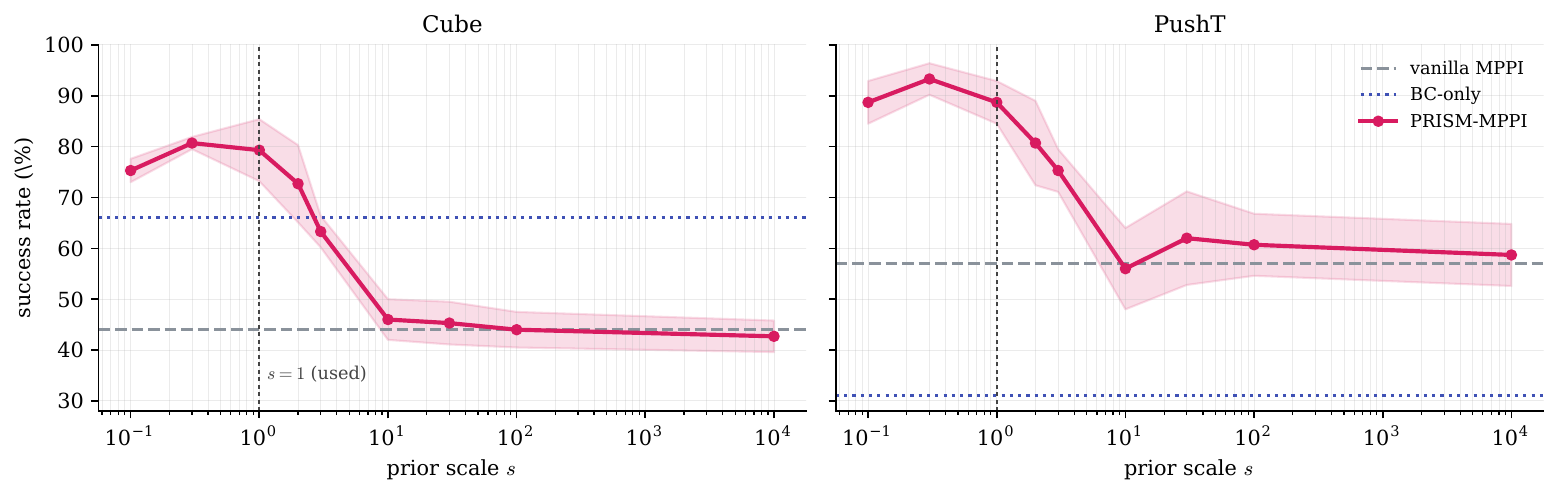}
\caption{\textbf{Prior-scale sweep.} SR against the prior scale $s$ (log axis), mean$\pm$std over three seeds. The dashed and dotted lines are vanilla MPPI and mean-only execution on the same task.}
\label{fig:asymptote}
\end{figure}

We vary the prior scale $s$ on PushT and Cube at $K=128$, using three seeds $\{0,1,42\}$ (Fig.~\ref{fig:asymptote}). Performance peaks around $s=0.3$ to $1$, supporting our default choice of $s=1$. Larger values reduce the prior's influence, and at $s=10^4$, success rates approach the prior-free results: $43\%$ versus $44\%$ on Cube and $59\%$ versus $57\%$ on PushT.

This trend is consistent with the PoG proposal approaching the default distribution as $s$ increases; in the limit the two planners are the same algorithm, since they now share every other setting.

\subsection{Real-Robot Evaluation}
\label{sec:exp-real}

\begin{figure}[t]
\centering
\includegraphics[width=\linewidth]{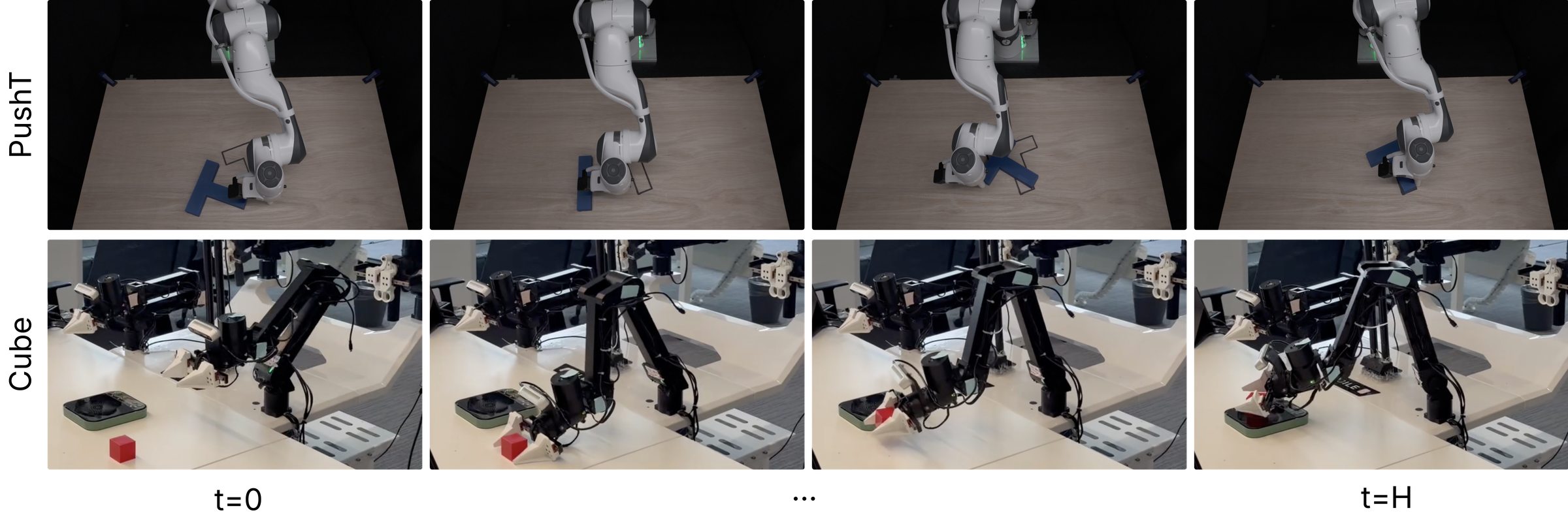}
\caption{\textbf{Real-robot rollouts.} Keyframes (left$\to$right) of one successful PRISM-MPPI episode per platform. \emph{Top:} Franka PushT. \emph{Bottom:} ARX~X5 cube manipulation.}
\label{fig:real-rollouts}
\end{figure}

We evaluate PRISM on two real-robot manipulation tasks: T-block pushing with a Franka Research 3 (FR3) arm and cube manipulation with an ARX X5 arm (Fig.~\ref{fig:real-rollouts}). Each platform uses a third-person RGB camera and a world model trained on task-specific teleoperation data. We compare planning with and without the learned prior over $40$ trials per condition, interleaved in randomized order, using $s=1$.

PRISM achieves higher observed success on both platforms: $14/40$ versus $10/40$ on Franka PushT, and $18/40$ versus $12/40$ on ARX cube manipulation. These results are consistent with the improvements observed in simulation, although more trials are needed to establish the size and reliability of the gains.

\begin{figure}[htbp]
\centering
\subfloat[Franka FR3 + PushT\label{fig:real-setup-franka}]{\includegraphics[width=0.49\linewidth]{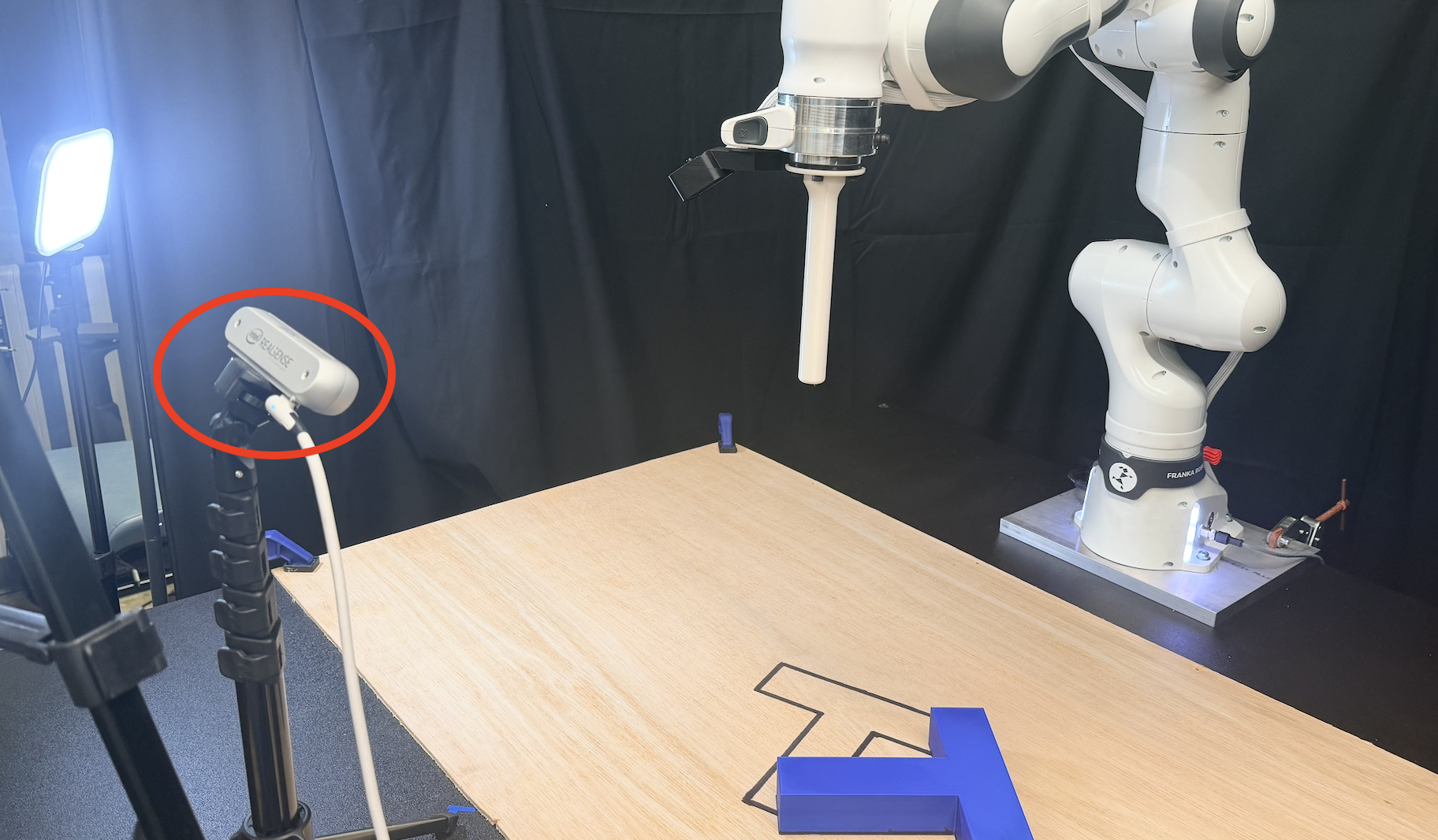}}
\hfill
\subfloat[ARX X5 + cube\label{fig:real-setup-arx}]{\includegraphics[width=0.49\linewidth]{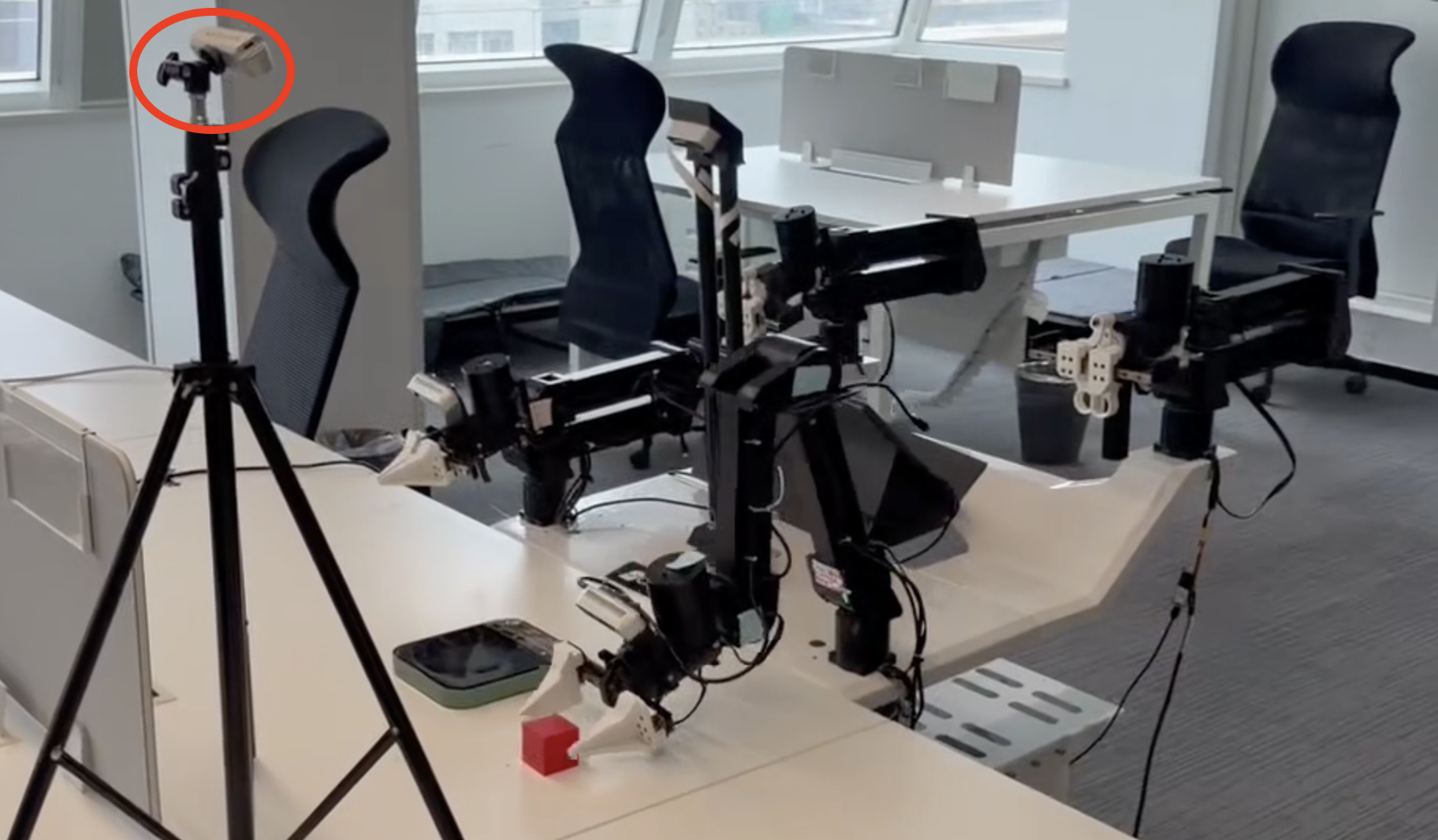}}
\caption{\textbf{Real-robot setups.} The red circle marks the tripod-mounted RealSense that is the planner's only RGB sensor; neither platform uses a wrist or egocentric camera. \emph{(a)} FR3, D455: the blue T-block must reach the marked target outline. \emph{(b)} ARX~X5 on an Agilex Mobile ALOHA, D435i: the arm transports a single red cube.}
\label{fig:real-setups}
\end{figure}

\parhead{Hardware and data.}
The FR3 setup uses a RealSense D455, Meta Quest 3 teleoperation, and an RTX 5080 for planning. The ARX X5 setup uses an Agilex Mobile ALOHA platform, a RealSense D435i, joystick teleoperation, and an RTX 4090. Neither setup uses wrist or egocentric cameras (Fig.~\ref{fig:real-setups}). The datasets contain $411$ episodes for Franka PushT and $201$ for ARX cube manipulation.

\parhead{Franka data and control protocol.}
The FR3 uses Cartesian-impedance control with planar end-effector displacements $(\Delta x,\Delta y)$; height, wrist orientation, and gripper are held fixed. RGB images are resized to $224\times224$. The Franka dataset contains $93{,}728$ frames recorded at $10$ Hz, comprising $200$ target-completion episodes and $211$ episodes ending at arbitrary positions. The world model uses all $411$ episodes, while the action head uses only the $200$ target-completion episodes with their terminal images as goals.

For this setup, planning uses $K{=}300$, $J{=}30$, $H{=}3$, and $B{=}5$, giving $1.5$ s of lookahead. Each replan starts from default scale $\sigma_\pi=1.0$, fuses the prior, and retains the resulting covariance during optimization. A goal image is captured once per session. The controller executes five denormalized actions at $10$ Hz before replanning, giving a $0.5$ s replanning interval. These control settings apply to the Franka setup.

\section{Limitations}
\label{sec:discussion}

Action predictability weakens for distant goals, and inaccurate, overconfident priors can impair planning. The utility of uncertainty-guided sampling depends on the task and the quality of the learned prior. PoG may discard multimodal structure, and future work will explore fusion methods that better preserve prior structure and adapt to prior reliability. Real-robot results use $40$ trials per condition and require broader validation.


\section{Conclusion}
\label{sec:conclusion}

We showed that representations learned to predict action consequences also support action intuition: goal-conditioned action priors can be learned from frozen world-model features through a lightweight readout. Action-prediction probes across four tasks, together with same-architecture random-initialization controls, provide evidence that predictive training yields features useful for action prediction. Through PRISM, we further demonstrated that the extracted priors improve planning performance in simulated and real-robot tasks without retraining the world model. Together, these findings reveal an underexplored capability of predictive representations: the same features can support both anticipating action consequences and proposing goal-directed actions.

\section*{Acknowledgments}
OpenAI's ChatGPT/Codex tools were used for drafting, restructuring, and language editing throughout, from author-provided descriptions and experimental results. The authors are responsible for the final content.

\bibliographystyle{IEEEtran}
\bibliography{references}

\end{document}